\documentclass[sigconf]{acmart}

\renewcommand\footnotetextcopyrightpermission[1]{}

\usepackage{booktabs}
\usepackage{tabularx}
\usepackage{amsmath}
\usepackage{graphicx}
\usepackage{algorithm}
\usepackage{algpseudocode}

\begin{document}

% \title{Comparing Post-Hoc and Co-Evolved Ensemble Strategies for Spiking Neural Networks}
\title{Co-Evolved Spiking Neural Network Ensembles via Marginal Contribution Fitness}

% ICONS 2026 conference metadata
\acmConference[ICONS '26]{International Conference on Neuromorphic Systems}
  {August 4--6, 2026}{Chicago, IL, USA}
\acmBooktitle{Proceedings of the International Conference on Neuromorphic Systems (ICONS '26),
  August 4--6, 2026, Chicago, IL, USA}
\acmYear{2026}
\copyrightyear{2026}

% Authors
\author{Catherine Rodriquez}
\email{crodr99@lsu.edu}
\affiliation{%
  \institution{Louisiana State University}
  \city{Baton Rouge}
  \state{Louisiana}
  \country{USA}
}

\author{James Ghawaly Jr.}
\email{jghawaly@lsu.edu}
\affiliation{%
  \institution{Louisiana State University}
  \city{Baton Rouge}
  \state{Louisiana}
  \country{USA}
}
% \author{Anonymous Author(s)}

\begin{abstract}
Evolutionary optimization of spiking neural networks (SNNs) becomes 
increasingly difficult as task complexity grows because they must 
search a combined topology--parameter space that grows 
super-exponentially with network size. We address this scaling 
challenge through a co-evolutionary ensemble framework in which a 
population of candidate SNNs is evolved with fitness defined by each 
network's marginal contribution to group performance. Grounded in 
cooperative game theory and difference evaluation functions from 
multiagent systems, this credit assignment rewards networks that 
consistently improve ensemble performance and penalizes redundancy, 
encouraging complementary specialization during evolution rather than 
relying on post-hoc combination of independently trained networks. We 
evaluate the approach on classification, regression, and control tasks under $\mu$Caspian neuromorphic hardware 
constraints. Co-evolved ensembles achieve statistically significant 
improvements over both single-network evolution and post-hoc 
ensembles across all tasks, with the most pronounced gains in 
control, where standard evolution fails to 
discover effective policies and co-evolution enables a qualitative 
transition to near-optimal performance.
\end{abstract}

% CCS Concepts (required by ACM; update to match your paper)


\maketitle

\section{Introduction}
Spiking neural networks (SNNs) offer significant advantages in energy efficiency and temporal processing, but training them effectively remains a major challenge. Evolutionary optimization approaches, such as Evolutionary Optimization for Neuromorphic Systems (EONS)~\cite{10.1145/3381755.3381758}, provide a gradient-free alternative that can simultaneously optimize both network topology and parameters. This flexibility makes evolutionary methods particularly well suited for neuromorphic systems, where spiking dynamics are non-differentiable and network structure is not fixed a priori. However, because EONS searches a combined topology--parameter space whose complexity grows super-exponentially with the number of neurons, the approach is fundamentally limited in the size and complexity of networks it can feasibly produce. As task complexity increases, standard evolution struggles to discover effective solutions within practical computational budgets.

One strategy for overcoming this scaling limitation is to decompose the problem into an ensemble of smaller networks that collectively solve the task. Ensemble learning has a long history in machine learning~\cite{10.5555/648054.743935, 10.1023/A:1018054314350, FREUND1997119}, and recent work has shown that evolving ensembles of SNNs using EONS can substantially outperform standard evolution on MNIST classification~\cite{elbrecht2020evolving} and binary classification in noisy multivariate time series data~\cite{Ghawaly_2025}. However, existing approaches construct ensembles sequentially or post-hoc, assembling independently trained networks into groups after evolution is complete. This leaves open the question of whether evolving networks cooperatively as ensemble members, rather than combining them after the fact, can yield further improvements through complementary specialization.

\subsection{Contributions}
In this work, we introduce a co-evolutionary ensemble framework for SNNs in which a single population of candidate networks is evolved with fitness determined by each network's marginal contribution to group performance. Drawing on cooperative game theory and difference evaluation functions from multiagent systems~\cite{shapley, wolpert2001optimal}, this formulation provides a principled credit assignment mechanism that rewards networks for consistently improving ensemble performance and penalizes redundancy. We evaluate the approach across three task families---classification, regression, and reinforcement learning control---and compare co-evolved ensembles against both single-network evolution and post-hoc ensemble construction.

\section{Related Work}

Evolutionary algorithms have long been explored as an alternative to gradient-based optimization for neural networks, particularly when differentiability is limited or architectural flexibility is desired. Early work demonstrated that evolution can optimize both network structure and parameters without backpropagation~\cite{6790655}, and methods such as NeuroEvolution of Augmenting Topologies (NEAT) established neuroevolution as a unified framework for evolving topology and weights~\cite{6790655}. More recent work has shown that genetic algorithms and evolution strategies can scale to modern deep learning and reinforcement learning problems~\cite{DBLP:journals/corr/abs-1712-06567, DBLP:journals/corr/abs-1802-01548}. However, these results rely on fixed architectures, where only connection weights are optimized. In contrast, evolving topology and weights together, as in EONS, introduces a combinatorial search over network structures that remains difficult to scale.

These properties are particularly well suited for SNNs, where discontinuous spike dynamics make gradient-based training challenging. Evolutionary approaches have been applied to SNN classification~\cite{pavlidis2005spiking} and control~\cite{qiu2018}, and frameworks such as EONS~\cite{10.1145/3381755.3381758} enable network structure and parameters to be evolved under neuromorphic hardware constraints. Although we use EONS through the TENNLab framework, the proposed methodology is applicable to a broad class of evolutionary algorithms that evolve both structure and parameters.

Ensemble methods have long been used to improve predictive performance and robustness. Classical approaches such as bagging and boosting reduce variance and improve generalization~\cite{10.1023/A:1018054314350, FREUND1997119}, while ensemble performance in neural networks is often driven by diversity among models~\cite{NIPS1994_b8c37e33}. Elbrecht et al.~\cite{elbrecht2020evolving} showed that sequentially evolving and combining SNNs with EONS can substantially outperform standard EONS on handwritten digit classification. Their boosting-inspired approach evolves networks one at a time and adds them to the ensemble based on examples misclassified by the current ensemble. This work demonstrated the value of ensemble learning for evolutionary SNN optimization while identifying co-evolutionary training strategies as an open direction.

The approach proposed in this work differs from prior ensemble methods by integrating ensemble formation directly into the evolutionary process through cooperative co-evolution. Rather than constructing ensembles sequentially or post-hoc, networks are evaluated collectively during training and assigned fitness based on their marginal contribution to group performance.

This framework draws on ideas from cooperative co-evolution and multiagent systems. Potter and De Jong~\cite{10.1007/3-540-58484-6_269} introduced cooperative co-evolution for real-valued function optimization, while Gomez et al.~\cite{JMLR:v9:gomez08a} demonstrated accelerated learning through cooperatively co-evolved synapses in fixed-topology recurrent networks. Neither approach considered SNNs, evolving network structure and parameters together, or neuromorphic hardware constraints. The marginal contribution fitness used here is grounded in the difference evaluation framework of Wolpert and Tumer~\cite{wolpert2001optimal}, which provides theoretical guarantees that the resulting credit assignment is both factored and aligned with the global objective.

By applying these principles to ensembles of topology-evolving SNNs, this work enables cooperative specialization during evolution, yielding more effective ensembles than post-hoc combination while using far fewer ensemble members than prior sequential approaches.

\section{General Methodology}
This work develops a generalizable framework for training and evaluating SNN ensembles using evolutionary optimization, focusing on cooperative behavior emerging through co-evolution. In this section, we first introduce the baseline approaches we compare to, then provide a detailed explanation of the co-evolved ensemble strategy.

\subsection{Standard EONS Baseline}
% original version used for submission:
% In the standard EONS training framework, a single SNN is returned from a population of candidate networks evolved over successive generations using evolutionary operators such as selection, crossover, and mutation~\cite{784219, 10.1145/3381755.3381758}. Each SNN in the population is evaluated independently according to a fitness function, and higher-performing networks are more likely to be selected for reproduction. This process iteratively improves the population by promoting networks that achieve better performance on the given task. 

% % This standard approach serves as the baseline in our experiments, where networks are optimized individually without considering interactions with other networks during training. Ensembles are formed only after evolution is complete, allowing us to compare post-hoc ensembling against the proposed co-evolved strategy.

% compressed version
In the standard EONS framework, candidate SNNs are evolved over successive generations using selection, crossover, and mutation~\cite{784219,10.1145/3381755.3381758}. Each network is evaluated independently according to a task-specific fitness function, and higher-performing networks are preferentially selected for reproduction. The highest-performing network obtained after evolution is returned as the final solution.

\subsection{Post-Evolved EONS Ensemble Baseline}
% orginal version used for submission
% In the post-evolved setting, networks are trained independently using the standard EONS framework described above, with no awareness of ensemble performance during evolution. After training is complete, ensembles are constructed by exhaustively evaluating all $\binom{P}{N}$ possible groups of size $N$ drawn from the final population of $P$ networks. Each candidate group is evaluated on the training data using the same task-specific aggregation function used in the co-evolved setting  and the group achieving the highest collective performance is selected as the post-evolved ensemble. 

% This exhaustive search identifies the best possible ensemble that can be formed from independently evolved networks and has proven effective in a variety of real-world applications~\cite{Ghawaly_2025}. 
% % Any performance advantage of co-evolution over this baseline cannot be attributed to the ensemble aggregation mechanism itself, since both strategies use identical aggregation functions. Instead, such an advantage would indicate that cooperative evaluation during training produces networks that are more complementary than those discovered by independent optimization followed by optimal post-hoc selection.

% updated version
In the post-evolved setting, candidate networks are evolved independently using the standard EONS framework, with fitness determined solely by individual task performance. As a result, networks are optimized without knowledge of the ensembles in which they may eventually participate. 

After evolution is complete, ensembles are constructed from the final population by examining all $\binom{P}{N}$ possible groups of size $N$, where $P$ is the population size. Each candidate group is evaluated on the training data using the same task-specific aggregation strategy employed by the co-evolved framework, ensuring that differences in performance are attributable to the training procedure rather than the prediction mechanism itself. The ensemble achieving the strongest collective performance is selected as the final post-evolved ensemble. 

This exhaustive search identifies the best ensemble that can be formed from independently evolved networks and has proven effective in real-world applications~\cite{Ghawaly_2025}. Unlike the proposed approach, however, any complementary behavior among ensemble members emerges only by chance rather than being explicitly encouraged throughout evolution.

\subsection{Co-Evolved Ensemble Strategy}
The proposed approach extends the standard EONS framework by evaluating candidate networks cooperatively rather than independently. During each generation, groups of size $N$ are formed from a population of candidate networks and evaluated according to their collective performance. Individual fitness is determined by each network's marginal contribution to the groups in which it participates, rewarding networks that consistently improve ensemble performance while penalizing redundant or detrimental behavior. This cooperative credit assignment encourages complementary specialization among ensemble members.

Let $G$ denote a group of $N$ networks containing network $i$, and let $F_{\text{group}}(G)$ denote the performance of that group. The fitness of network $i$ is defined as its average marginal contribution across all groups containing it:

\begin{equation}
F_i = \frac{1}{|S_i|}
\sum_{G \in S_i}
\left[
F_{\text{group}}(G)
-
F_{\text{group}}(G \setminus \{i\})
\right]
\end{equation}

where $S_i$ is the set of groups containing network $i$. For a population of size $P$, each network participates in $\binom{P-1}{N-1}$ possible groups. Networks that provide little or no benefit receive near-zero or negative marginal fitness, creating selective pressure toward non-redundant ensemble members.

\begin{algorithm}[t]
\caption{Co-Evolved EONS}
\begin{algorithmic}[1]
\State Initialize an EONS population of $P$ candidate SNNs
\For{each generation}
    \State Form candidate ensembles of size $N$
    \For{each ensemble $G$}
        \State Evaluate group performance $F_{\text{group}}(G)$
        \For{each member $i \in G$}
            \State $\Delta_i \leftarrow F_{\text{group}}(G)-F_{\text{group}}(G\setminus\{i\})$
            \State Accumulate $\Delta_i$ into network $i$'s fitness
        \EndFor
    \EndFor
    \State Average each network's accumulated marginal contribution
    \State Apply standard EONS selection, crossover, and mutation
\EndFor
\State Return the best ensemble discovered
\end{algorithmic}
\end{algorithm}

This formulation is a fixed-coalition-size variant of the marginal contribution principle from cooperative game theory~\cite{shapley} and is closely related to difference evaluation functions from multiagent systems~\cite{wolpert2001optimal}, which Wolpert and Tumer showed to be both \emph{factored} and aligned with the global objective. Across all experiments, the definition of $F_{\text{group}}$ is adapted to the task: classification accuracy for supervised classification, error reduction for regression, and cumulative reward for reinforcement learning. In all cases, evolution is guided toward networks that work effectively as part of a group rather than optimizing purely for individual performance~\cite{JMLR:v9:gomez08a,10.1007/3-540-58484-6_269}.

\subsection{Training}
% needs work
All networks are trained with EONS in the TENNLab framework using the hyperparameters outlined in Table~\ref{tab:hyperparams}, targeting the $\mu$Caspian hardware platform~\cite{mitchell2020caspian}. The post-evolved and co-evolved strategies used a modified version of the EONS driver.
% Training is performed on virtualized hardware representations provided through TENNLab, allowing efficient exploration of network architectures prior to deployment on neuromorphic harware systems. 
The task specific fitness definitions and evaluation procedures for each experiment are detailed in Section~\ref{sec:experiments}. Detailed descriptions of these parameters are provided in prior work and the EONS documentation \cite{10.1145/3381755.3381758}.

% \begin{table}[t]
% \caption{EONS Hyperparameters within TENNLab}
% \label{tab:hyperparams}
% \centering
% \begin{tabular}{ll}
% \toprule
% Hyperparameter & Value \\
% \midrule
% population\_size & 100 \\
% num\_generations & 100 \\
% starting\_nodes & 50 \\
% starting\_edges & 50 \\
% merge\_rate & 0 \\
% multi\_edges & 0 \\
% mutation\_rate & 0.9 \\
% number\_mutations & 4 \\
% random\_factor & 0.05 \\
% crossover\_rate & 0.5 \\
% selection\_type & tournament \\
% tournament\_size\_factor & 0.9 \\
% node\_mutations & \{Threshold: 1.0\} \\
% edge\_mutations & \{Weight: 0.7, Delay: 0.3\} \\
% num\_best & 3 \\
% add\_node\_rate & 0.5 \\
% delete\_node\_rate & 0.25 \\
% add\_edge\_rate & 0.75 \\
% delete\_edge\_rate & 0.25 \\
% node\_params\_rate & 2.5 \\
% edge\_params\_rate & 2.5 \\
% net\_params\_rate & 0 \\
% \bottomrule
% \end{tabular}
% \end{table}
\begin{table}[t]
\caption{EONS Hyperparameters within TENNLab}
\label{tab:hyperparams}
\centering
\setlength{\tabcolsep}{3pt}
\small
\begin{tabular}{llll}
\toprule
Hyperparameter & Value & Hyperparameter & Value \\
\midrule
population\_size & 100 & selection\_type & tournament \\
num\_generations & 100 & tournament\_size\_factor & 0.9 \\
starting\_nodes & 50 & num\_best & 3 \\
starting\_edges & 50 & add\_node\_rate & 0.5 \\
mutation\_rate & 0.9 & delete\_node\_rate & 0.25 \\
number\_mutations & 4 & add\_edge\_rate & 0.75 \\
crossover\_rate & 0.5 & delete\_edge\_rate & 0.25 \\
random\_factor & 0.05 & node\_params\_rate & 2.5 \\
merge\_rate & 0 & edge\_params\_rate & 2.5 \\
multi\_edges & 0 & net\_params\_rate & 0 \\
node\_mutations & thresh: 1.0 & edge\_mutations & W: 0.7, D: 0.3 \\
\bottomrule
\end{tabular}
\end{table}
\subsection{Motivation: Search Space Complexity}

Unlike algorithms that optimize parameters over a fixed architecture, EONS simultaneously searches over network topologies and discrete hardware-constrained parameters.
We formalize this search space to motivate the ensemble decomposition as a strategy for scaling evolutionary neuromorphic optimization to problems requiring larger networks.

Let the target hardware be characterized by a maximum neuron count $n_{\max}$, a maximum synapse count $m_{\max}$, a per-neuron parameter bit-width $b_v$, and a per-synapse parameter bit-width $b_e$.
A candidate network is a directed graph $G = (V, E)$ with $|V| = n \leq n_{\max}$ neurons and $|E| = m \leq m_{\max}$ synapses, where $E \subseteq V \times V$ may include self-loops.
For $n$ neurons, there are $n^2$ potential directed edges.
Each slot is independently either absent (one state) or present with a hardware-constrained parameter assignment ($2^{b_e}$ states), giving $1 + 2^{b_e}$ states per slot.
Including $2^{b_v}$ parameter configurations per neuron, the total number of distinct candidate networks of size $n$ is:
\begin{equation}
    |\Omega(n)| \;=\; 2^{\,b_v\, n}\;\bigl(1 + 2^{b_e}\bigr)^{n^2}
    \label{eq:omega}
\end{equation}
Defining $\beta = \log_2(1 + 2^{b_e}) \approx b_e$ for $b_e \geq 4$:
\begin{equation}
    \log_2|\Omega(n)| \;=\; b_v\,n \;+\; \beta\,n^2
    \label{eq:log_omega}
\end{equation}
The dominant $\beta\, n^2$ term reflects the $n^2$ potential synaptic connections, each carrying both a topological binary choice and a discrete parameter assignment.
The search space grows super-exponentially with the number of neurons.
When $n > \lfloor\sqrt{m_{\max}}\rfloor$, the hardware synapse limit caps the edge count below $n^2$, reducing the effective search space; however, the growth remains super-exponential due to the combinatorial cost of selecting which $m_{\max}$ edges to instantiate from the $n^2$ available slots.

Now consider an ensemble of $N$ sub-networks, each evolved in its own population under the same hardware constraints with no additional size restrictions.
Let $n_{\text{mono}}$ denote the characteristic neuron count of networks produced by single-population evolution and $\bar{n}$ the characteristic count for co-evolved sub-networks.
Because the $N$ populations search independently (coupled only through the co-evolutionary fitness signal), the total ensemble search space is:
\begin{equation}
    \log_2|\Omega_{\text{ens}}| \;=\; N\bigl[b_v\,\bar{n} + \beta\,\bar{n}^2\bigr]
    \label{eq:log_omega_ens}
\end{equation}
For networks large enough that the quadratic term dominates ($n \gg b_v / \beta$), the ratio of log-sizes is:
\begin{equation}
    R \;\equiv\; \frac{\log_2|\Omega_{\text{ens}}|}{\log_2|\Omega_{\text{mono}}|}
    \;\approx\; \frac{N\,\bar{n}^{\,2}}{n_{\text{mono}}^{\,2}}
    \label{eq:ratio}
\end{equation}
Since these are log-sizes, $R < 1$ corresponds to $|\Omega_{\text{ens}}| \approx |\Omega_{\text{mono}}|^R$.
The ensemble search space is smaller whenever:
\begin{equation}
    \bar{n} \;<\; \frac{n_{\text{mono}}}{\sqrt{N}}
    \label{eq:critical_condition}
\end{equation}
For $N = 3$, this requires co-evolved sub-networks smaller than approximately $57.7\%$ of the monolithic network size.
Co-evolutionary fitness based on marginal contribution (Equation~1) creates selective pressure toward smaller, specialized networks by rewarding complementary behavior rather than individual performance, making this condition plausible in practice.

The reduction in search space also improves the effectiveness of mutation-based search.
The 1-mutation neighborhood of a network grows polynomially with $n$ (bounded by $O(n^2 \cdot 2^{b_e})$), while the search space grows super-exponentially.
The fraction of the space reachable in a single mutation therefore satisfies $\alpha(\bar{n}) / \alpha(n_{\text{mono}}) \sim 2^{\,\beta(n_{\text{mono}}^2 - \bar{n}^{\,2})}$, meaning each mutation in a sub-network explores an exponentially larger fraction of the relevant space.

The cost of this decomposition is a restriction in representational capacity: the ensemble can only express functions of the form $g(f_1(x), \ldots, f_N(x))$ where $g$ is the fixed aggregation and each $f_i$ is independently computable by a sub-network.
When this decomposability assumption holds, the exponential search space reduction and improved mutation reachability enable evolutionary optimization to scale to problems that are infeasible for standard evolution under the same computational budget.

\section{Experiments}\label{sec:experiments}
We present several application case studies to evaluate the proposed co-evolved ensemble approach. These experiments allow us to observe how cooperative evolutionary training behaves across different problem types while maintaining a consistent training framework. Unless otherwise noted, all experiments use the default EONS hyperparameters shown in Table~\ref{tab:hyperparams}.

 \subsection{Supervised Classification Tasks}

\paragraph{1) Datasets:} We use three benchmark datasets from the Scikit-learn library~\cite{sklearn}: Wine, Iris, and Breast Cancer. For each dataset, we use a 2/3 training and 1/3 testing split, kept consistent across all methods to ensure a fair comparison. Class predictions are determined using a rate-normalized argmax approach, where the total spike count for each output class is divided by the total number of spikes produced across all output neurons, and the class with the highest normalized value is selected as the predicted label~\cite{d7dd678d2cc14f6eaaa4a1727d3e1c9c}.

\paragraph{2) Network Ensembling:} We use an aggregation approach based on the combined outputs of the networks. For a given input $x$, each network produces a vector of class scores, and these outputs are summed across the ensemble. The final prediction is determined by selecting the class with the highest combined score:

\begin{equation}
\hat{y} = \arg\max_{c} \sum_{i=1}^{N} f_i(x)_c
\end{equation}
where $f_i(x)_c$ represents the score assigned to class $c$ by the $i$-th network, and $N$ is the number of networks in the ensemble. This approach allows networks that may not perform optimally on their own to contribute meaningfully when combined with others.

\paragraph{3) Co-Evolved Fitness Function:} The cooperative fitness formulation for classification instantiates the general marginal contribution principle (Equation~1) using classification accuracy of the aggregated ensemble outputs as the group performance measure $F_{\text{group}}$. For a group $G$ of $N$ networks and a training set of $M$ samples, the group performance is:

\begin{equation}
F_{\text{group}}(G) = \frac{1}{M} \sum_{m=1}^{M} \mathbf{1}\!\left[\arg\max_{c} \sum_{i \in G} f_i(x_m)_c = y_m\right]
\end{equation}

where $y_m$ is the true label for sample $x_m$. The fitness of each network is then computed as its average marginal contribution to this accuracy across all partner combinations, as defined in Equation~1. A network that improves ensemble accuracy across many partner combinations receives high fitness, while one that is redundant with its partners receives low fitness regardless of its individual accuracy.

\paragraph{4) Experimental Protocol:} For each dataset and method, we perform 30 runs using different random seeds. The standard EONS baseline and post-evolved ensemble baseline follow the procedures described in Sections~3.1 and~3.2, respectively. Co-evolved ensembles are trained using the same random seeds to enable direct paired comparison, with ensemble accuracy on the training data determining group fitness during evolution. After training, all methods are evaluated on the held-out test set.

\begin{figure*}[t]
\centering
\includegraphics[width=\textwidth]{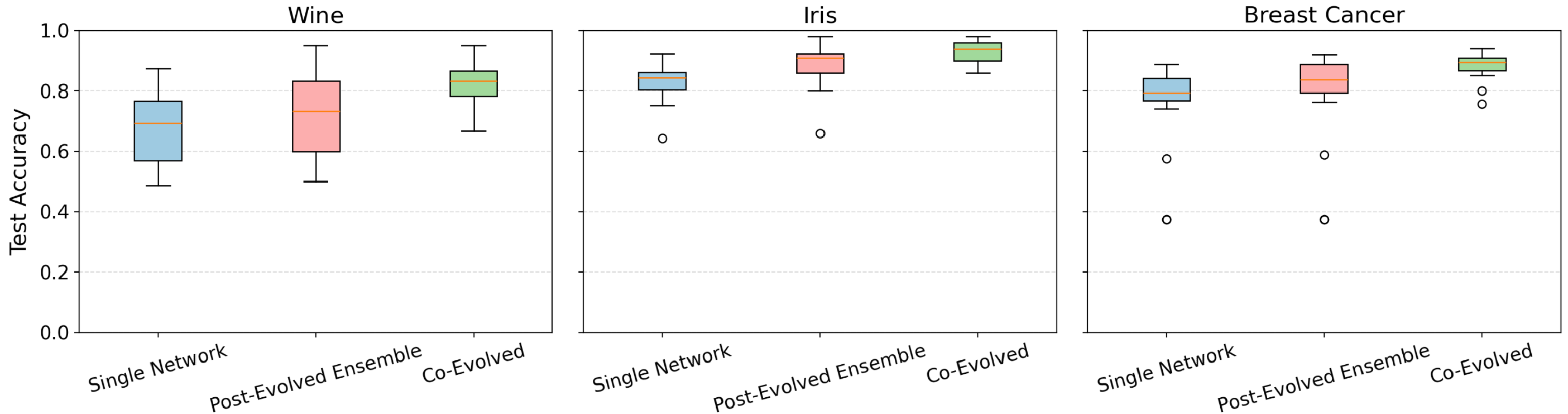}
\caption{Comparison of classification results.}
\label{fig:classification_results}
\end{figure*}

\subsection{Regression Tasks}

\paragraph{1) Datasets:}
We evaluate the co-evolved ensemble approach on four regression benchmarks: California Housing and Diabetes from the Scikit-learn library~\cite{sklearn}, and the Concrete Compressive Strength and Superconductivity datasets from the UCI Machine Learning Repository~\cite{concrete_compressive_strength_165, superconductivty_data_464}. The Superconductivity dataset introduces a more highly nonlinear regression setting compared to the others. As with the classification experiments, each dataset is split into 2/3 training and 1/3 testing partitions, and the same splits are used across all methods to ensure a fair comparison.

\paragraph{2) Network Ensembling:}
For regression tasks, the ensemble prediction is computed as the average of the individual network outputs (rate decoded). Given an input $x$, each network $f_i$ produces a scalar prediction, and the ensemble output is:

\begin{equation}
\hat{y} = \frac{1}{N} \sum_{i=1}^{N} f_i(x)
\end{equation}

where $N$ is the number of networks in the ensemble. This averaging formulation naturally reduces variance in predictions and provides a smooth mechanism for combining networks with complementary strengths.

\paragraph{3) Co-Evolved Fitness Function:}
The cooperative fitness formulation for regression instantiates the general marginal contribution principle (Equation~1) using prediction error as the group performance measure. Let $E_{\text{group}}(i,j,k)$ denote the mean absolute error (MAE) of the ensemble when all three networks contribute to the averaged prediction, and let $E_{\text{group}}(j,k)$ denote the error when network $i$ is excluded. The fitness of network $i$ is its average marginal error reduction across all partner combinations:

\begin{equation}
F_i = \frac{1}{|S_i|} \sum_{(j,k) \in S_i} \left[ E_{\text{group}}(j, k) - E_{\text{group}}(i, j, k) \right]
\end{equation}

where $S_i$ is the set of all distinct partner pairs drawn from the population, excluding network $i$. The subtraction order is reversed relative to the classification case: because lower error is better, a positive fitness value indicates that including network $i$ reduced the group's error. A network that merely duplicates the predictions of its partners contributes little to the average and receives a low fitness score, since its inclusion barely changes the group error. The evolutionary pressure therefore naturally encourages diversity and specialization, favoring networks that are accurate in regions of the input space where the others are not.

\paragraph{4) Experimental Protocol:}
For each dataset, we perform 75 runs using different random seeds. The standard EONS baseline and post-evolved ensemble baseline follow the procedures described in Sections~3.1 and~3.2, respectively. Co-evolved ensembles are trained using the same random seeds to enable direct paired comparison, with ensemble MAE on the training data determining group fitness during evolution. After training, all methods are evaluated on the held-out test set.

\begin{figure}[t]
\centering
\includegraphics[width=\columnwidth]{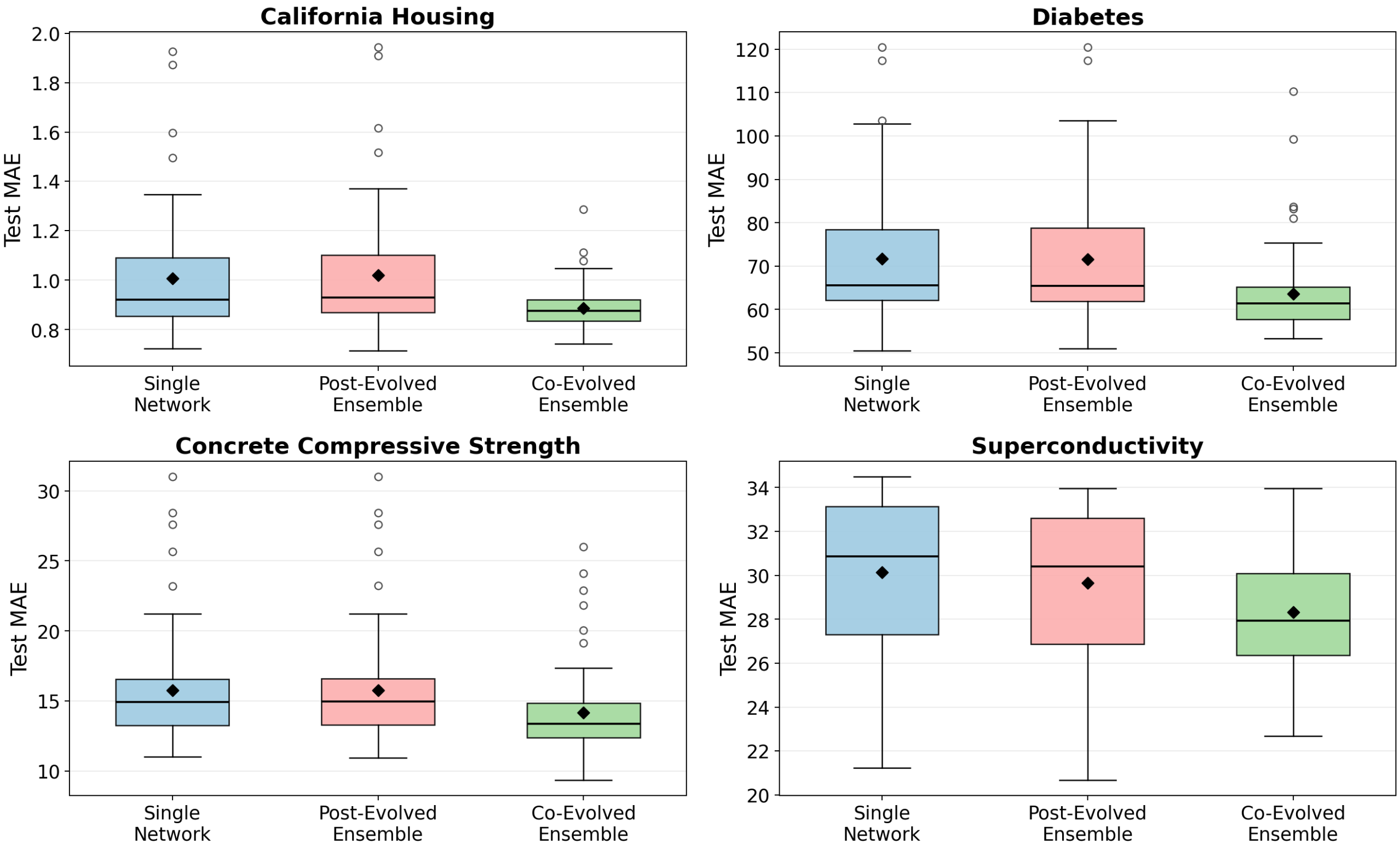}
\caption{Comparison of regression results.}
\label{fig:regression_results}
\end{figure}

\subsection{Control Tasks}

\paragraph{1) Environment:} We evaluate the co-evolved ensemble approach on the CartPole-v1 benchmark from the OpenAI Gym framework~\cite{brockman2016openaigym}, where the objective is to balance a pole for up to 500 timesteps per episode. During evolutionary training, fitness is computed as the average episodic return over 5 episodes. Final evaluation is conducted over 100 episodes.

\paragraph{2) Network Ensembling:} Environment observations are first converted into spike-based representations, enabling the neuromorphic networks to process continuous state inputs. At each timestep, each network produces action-value estimates for all available actions. Following prior work on ensemble reinforcement learning~\cite{DBLP:journals/corr/OsbandBPR16, DBLP:journals/corr/abs-2007-04938}, these estimates are averaged across the ensemble and the action with the highest averaged value is selected:

\begin{equation}
Q_{\text{ens}}(s_t, a) = \frac{1}{N} \sum_{i=1}^{N} Q_i(s_t, a)
\end{equation}

\begin{equation}
a_t = \arg\max_{a \in \mathcal{A}} Q_{\text{ens}}(s_t, a)
\end{equation}

This formulation enables multiple networks to collaboratively influence action selection at each timestep, allowing the ensemble to leverage diverse value estimates for sequential decision-making.

\paragraph{3) Co-Evolved Fitness Function:}
The cooperative fitness formulation for control instantiates the general marginal contribution principle (Equation~1) using cumulative episodic reward as the group performance measure, modified to account for the computational cost of episode-based evaluation. Let $R_{\text{group}}(i,j,k)$ denote the average cumulative reward when networks $i$, $j$, and $k$ act as an ensemble, and $R_{\text{group}}(j,k)$ the average reward when network $i$ is excluded.

In the classification and regression settings, every candidate network is evaluated across all $\binom{P-1}{N-1}$ partner combinations in every generation. For control tasks, this exhaustive evaluation is computationally prohibitive because each evaluation requires running full episodes rather than a single forward pass over a dataset. We therefore adopt a two-phase approach. In the first generation ($g = 1$), fitness is computed using exhaustive evaluation:

\begin{equation}
F_i^{(1)} = \frac{1}{|S_i|} \sum_{(j,k) \in S_i} \left[ R_{\text{group}}(i,j,k) - R_{\text{group}}(j,k) \right]
\end{equation}

Each network is then assigned to the group in which it achieved its highest marginal contribution:

\begin{equation}
(j^*, k^*) = \arg\max_{(j,k) \in S_i} \left[ R_{\text{group}}(i,j,k) - R_{\text{group}}(j,k) \right]
\end{equation}

For all subsequent generations ($g > 1$), group assignments are held fixed:

\begin{equation}
F_i^{(g)} = R_{\text{group}}(i, j^*, k^*) - R_{\text{group}}(j^*, k^*), \quad g > 1
\end{equation}

This reduces the per-generation cost to a single set of episode rollouts per group. While fixing groups after the first generation sacrifices the ability to reassign partnerships as networks evolve, the initial exhaustive search ensures that trios are formed based on measured complementarity rather than random assignment. A positive fitness value indicates that network $i$ improved the ensemble's performance, but here the formulation rewards networks whose action-value estimates complement those of their partners over the course of an episode rather than on independent samples.

\paragraph{4) Experimental Protocol:}
We perform 50 runs with a fixed random seed of 0 for environment initialization. The standard EONS baseline and post-evolved ensemble baseline follow the procedures described in Sections~3.1 and~3.2, respectively. Co-evolved ensembles are trained under the same conditions. After training, all methods are evaluated over 100 test episodes for comparison.

\begin{figure}[t]
\centering
\includegraphics[width=\columnwidth]{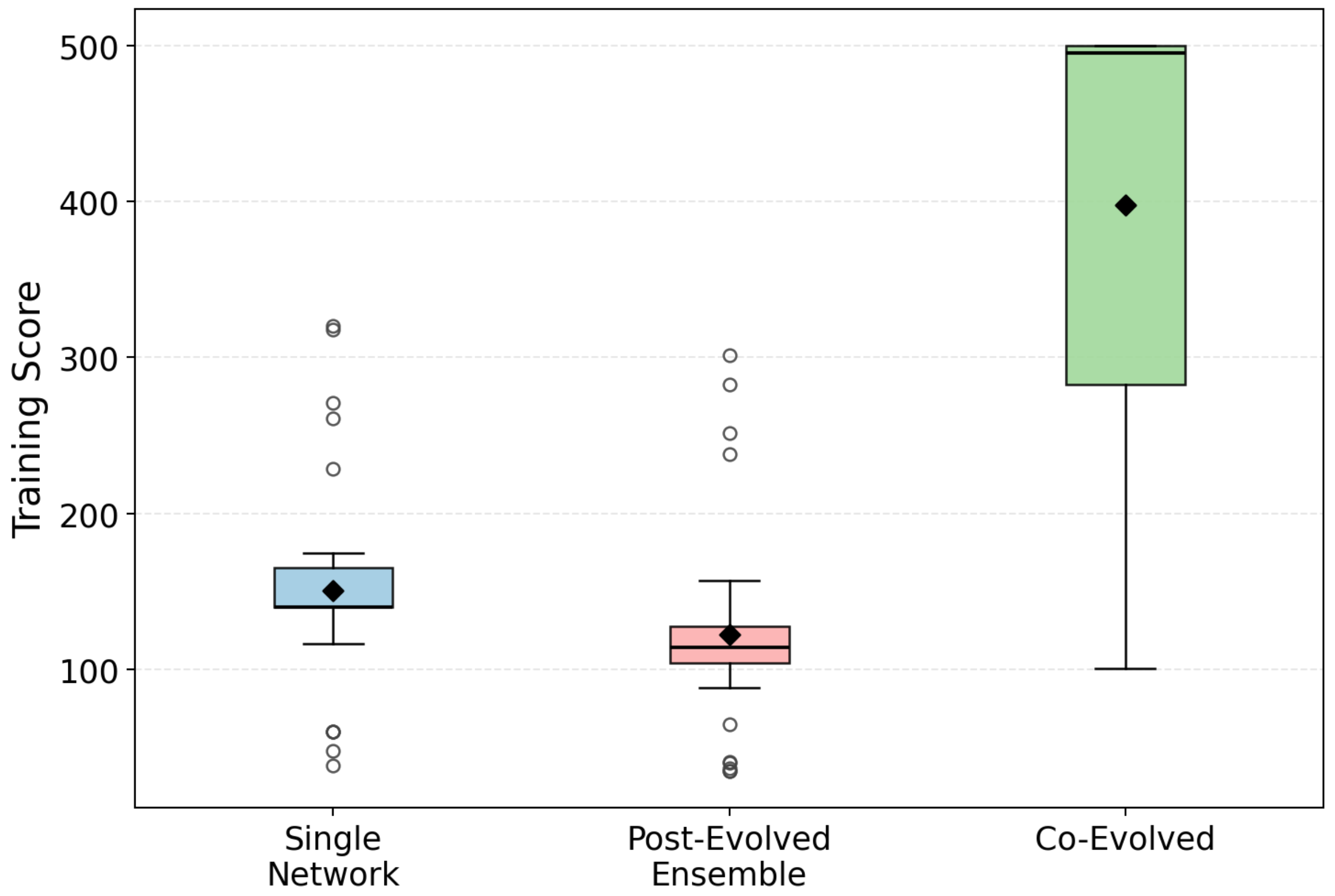}
\caption{Comparison of control results.}
\label{fig:control_results}
\end{figure}
% =====================================================
% scaling
 
\subsection{Scaling Studies}
% In the experiments described above, we primarily evaluate ensemble sizes of three as this is the maximum number of networks that can typically fit on Caspian and has proven to work well on real-world tasks with post-evolutionary ensembling~\cite{Ghawaly_2025}. However, to examine how the co-evolutionary framework behaves as the number of interacting populations increases, we conduct ensemble scaling studies on both the Wine classification dataset and the Superconductivity regression dataset. In these experiments, we vary the size of the ensembles from two to ten for both datasets.
% % , extending beyond the standard three-network ensemble configuration used in the primary experiments.

% For each setting, networks are trained using the same cooperative procedure as described above, with group formation adjusted to match the number of networks in each ensemble. Performance is evaluated using the same task-specific metrics as in the corresponding experiments. 
% % These studies are designed to assess how increasing the number of co-evolved populations influences performance and whether larger ensembles continue to provide meaningful improvements.

In the experiments described above, we primarily evaluate ensemble sizes of three as this is the maximum number of networks that can typically fit on Caspian and has proven to work well on real-world tasks with post-evolutionary ensembling~\cite{Ghawaly_2025}. To study scalability, we vary ensemble size from two to ten on the Wine classification and Superconductivity regression datasets. For each setting, networks are trained using the same cooperative procedure with group formation adjusted to the ensemble size, and evaluated using the same task-specific metrics as in the corresponding experiments.

% classification
\begin{figure}[t]
\centering
\includegraphics[width=\columnwidth]{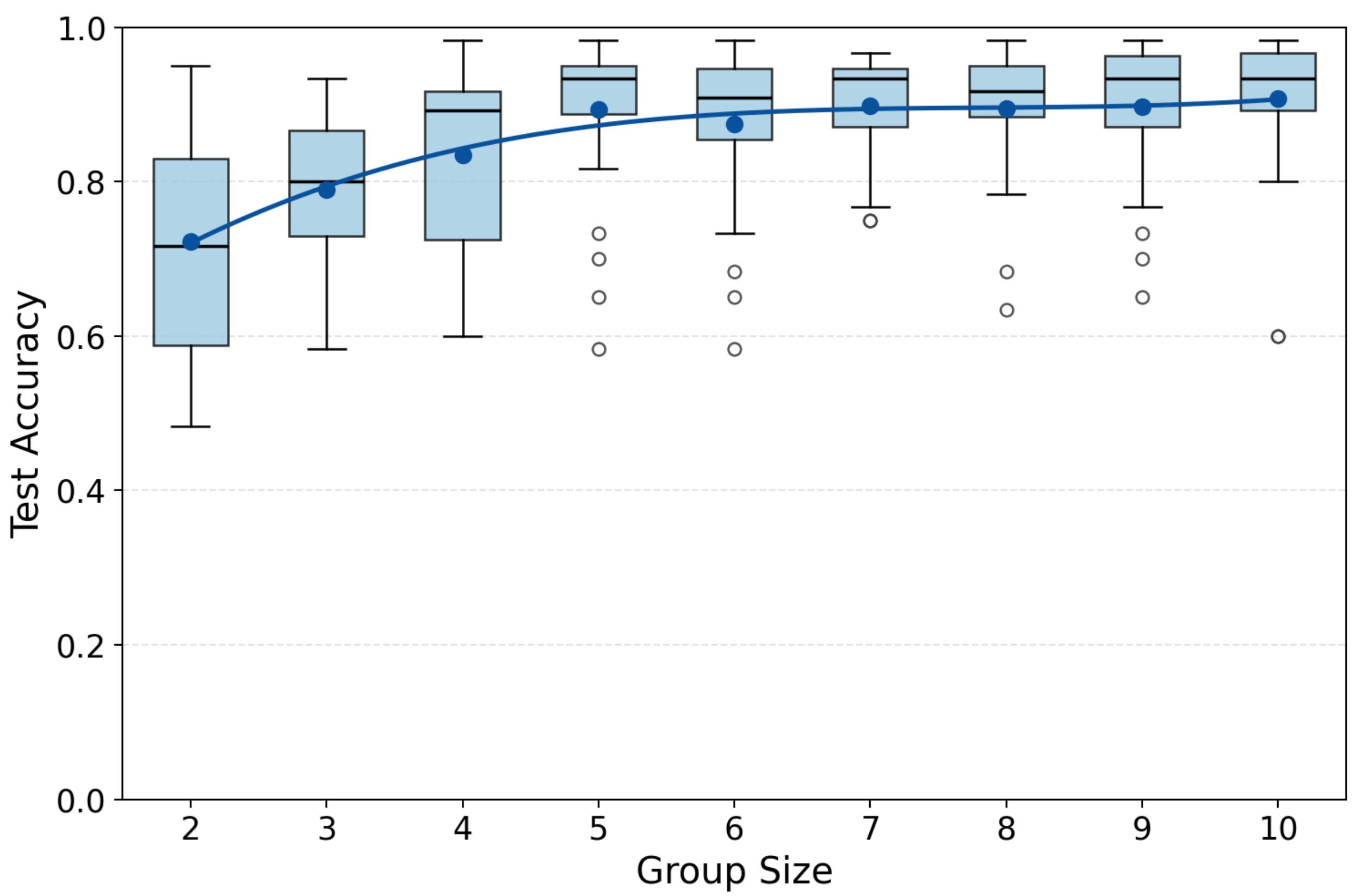}
\caption{Comparison of scaling different group sizes for co-evolution on the wine classification dataset.}
\label{fig:wine_coevolved_scaling}
\end{figure}

% regression
\begin{figure}[t]
\centering
\includegraphics[width=\columnwidth]{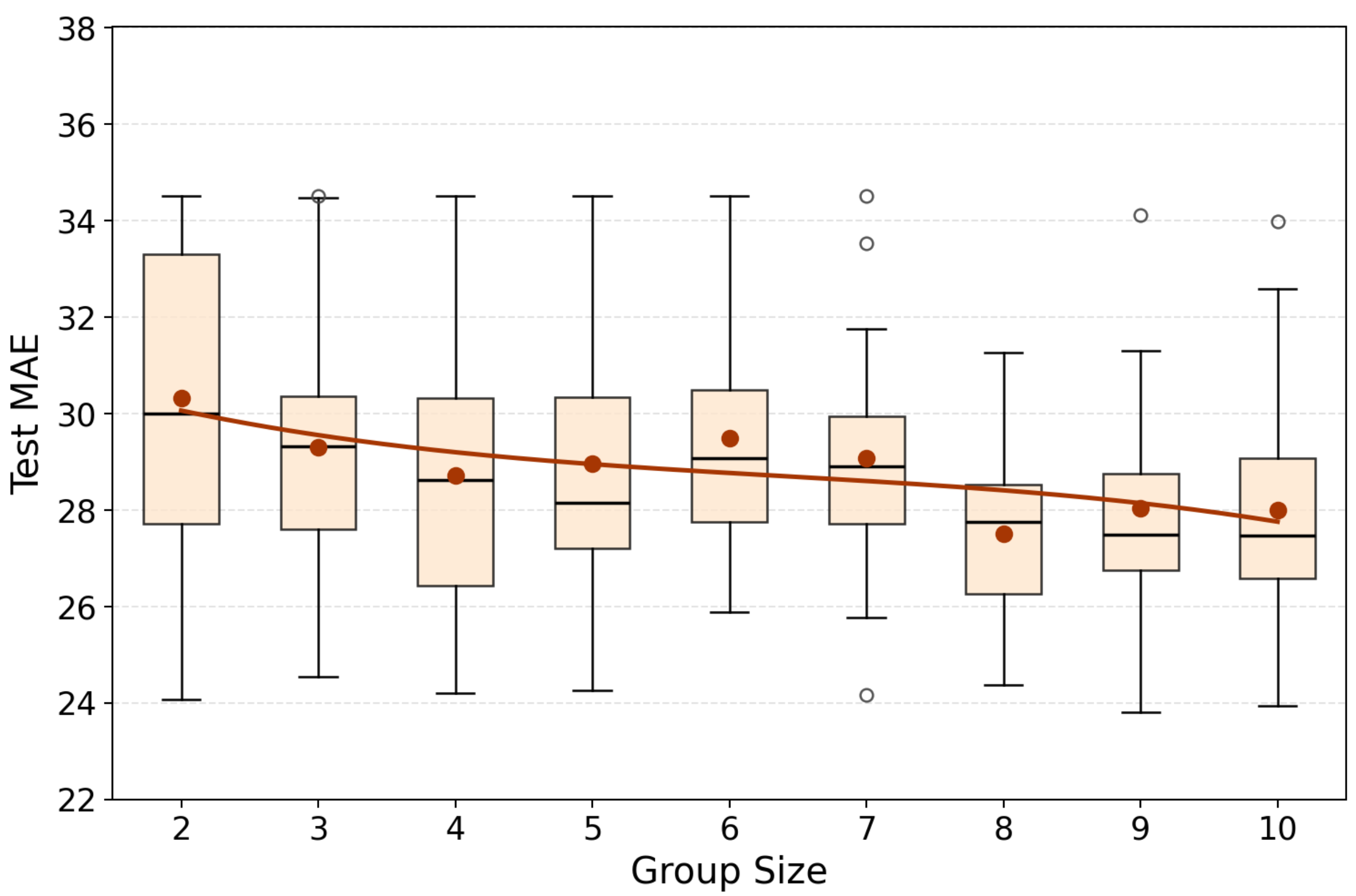}
\caption{Comparison of scaling different group sizes for co-evolution on the superconductivity dataset.}
\label{fig:superconductivity_coevolved_scaling}
\end{figure}

\section{Discussion}
% original version used for submission
% To assess whether the observed performance differences are statistically significant, we conduct one-sided Wilcoxon signed-rank tests comparing co-evolved ensembles against both single networks and post-evolved ensembles across all datasets. The results, shown in Table~\ref{tab:wilcoxon}, indicate statistically significant improvements across all classification, regression, and control tasks. Strong significance is observed consistently across datasets, confirming that the gains from co-evolution are not driven by random variation but reflect reliable performance improvements across runs.

% updated version compressed for page limit concerns
To assess statistical significance, we conduct one-sided Wilcoxon signed-rank tests comparing co-evolved ensembles against both single networks and post-evolved ensembles across all datasets. The results in Table~\ref{tab:wilcoxon} show statistically significant improvements across classification, regression, and control tasks, indicating that the gains from co-evolution are consistent across runs rather than due to random variation.

\begin{figure*}[t]
\centering
\includegraphics[width=0.8\textwidth]{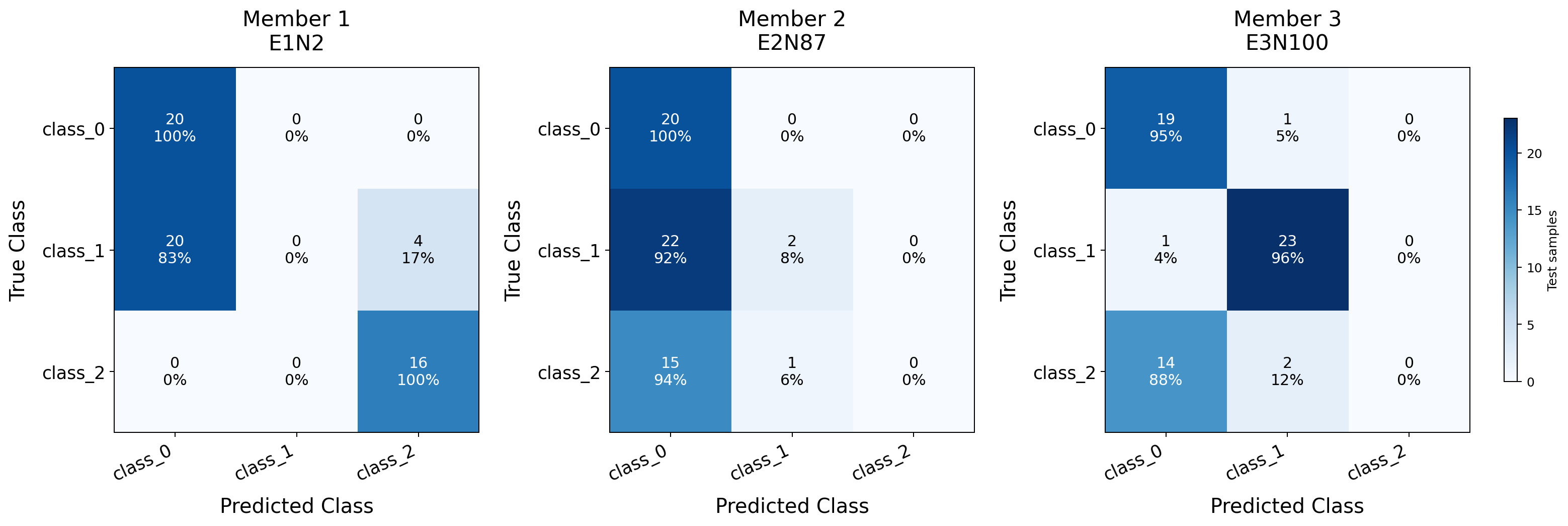}
\caption{Confusion matrices for the best-performing co-evolved ensemble on the Wine testing dataset. Individual members specialize on different classes, demonstrating complementary behavior that contributes to improved ensemble accuracy.}
\label{fig:member_specialization}
\end{figure*}

\begin{table}[t]
\caption{Wilcoxon signed-rank test results comparing co-evolved ensembles to baseline methods}
\label{tab:wilcoxon}
\centering
\setlength{\tabcolsep}{4pt}
\begin{tabular}{lcc}
\toprule
Dataset & Co vs Single (p) & Co vs Post-Evolved (p) \\
\midrule
\multicolumn{3}{c}{\textit{Classification (Accuracy $\uparrow$)}} \\
Wine & $9.15\text{e-}06$ & $1.95\text{e-}03$ \\
Iris & $2.98\text{e-}07$ & $7.17\text{e-}03$ \\
Breast Cancer & $1.49\text{e-}07$ & $7.41\text{e-}04$ \\
\midrule
\multicolumn{3}{c}{\textit{Regression (MAE $\downarrow$)}} \\
Housing & $1.16\text{e-}04$ & $2.34\text{e-}05$ \\
Diabetes & $5.55\text{e-}05$ & $8.33\text{e-}05$ \\
Concrete & $9.69\text{e-}04$ & $1.16\text{e-}03$ \\
Supercond. & $7.39\text{e-}04$ & $5.95\text{e-}03$ \\
\midrule
\multicolumn{3}{c}{\textit{Control (Reward $\uparrow$)}} \\
CartPole & $7.82\text{e-}07$ & $5.11\text{e-}07$ \\
\bottomrule
\end{tabular}
\end{table}

\paragraph{A. Classification Tasks:}\mbox{}\\

% original version used for submission:
% Across all three datasets, co-evolved ensembles achieve higher test accuracy than single networks and post-evolved ensembles, as shown in Figure~\ref{fig:classification_results}. The improvement is most pronounced on Wine and Breast Cancer, where the separation between methods is clearly visible. In addition to higher median accuracy, the co-evolved approach produces tighter distributions, indicating more stability across runs. These trends are further supported by the summary statistics in Table~\ref{tab:classification_table_results}, where co-evolved ensembles achieve higher average accuracy along with reduced standard deviation compared to the baseline approaches. Post-evolved ensembling provides some improvement over single networks, but the gains are less consistent and do not match the performance of co-evolved training.

% updated version
Across all three datasets, co-evolved ensembles achieve higher test accuracy than both single networks and post-evolved ensembles, as shown in Figure~\ref{fig:classification_results}. The gains are particularly pronounced on Wine and Breast Cancer, and the tighter distributions indicate improved stability across runs. These trends are reflected in the summary statistics of Table~\ref{tab:classification_table_results}. While post-evolved ensembling provides some improvement over single networks, the gains are smaller and less consistent than those achieved through cooperative training. 

\begin{table}[t]
\caption{Classification Performance (Accuracy $\uparrow$)}
\label{tab:classification_table_results}
\centering
\begin{tabular}{lccc}
\toprule
\textbf{Dataset} & \textbf{Single Net.} & \textbf{Post-Evolved} & \textbf{Co-Evolved} \\
 & ($\mu \pm \sigma$) & ($\mu \pm \sigma$) & ($\mu \pm \sigma$) \\
\midrule
Wine & 0.674 $\pm$ 0.115 & 0.729 $\pm$ 0.133 & \textbf{0.831 $\pm$ 0.067} \\
Iris & 0.828 $\pm$ 0.061 & 0.890 $\pm$ 0.068 & \textbf{0.931 $\pm$ 0.041} \\
Breast Cancer & 0.751 $\pm$ 0.155 & 0.784 $\pm$ 0.169 & \textbf{0.884 $\pm$ 0.039} \\
\bottomrule
\end{tabular}
\end{table}

To better understand the source of these improvements, Figure~\ref{fig:member_specialization} shows confusion matrices for the best-performing co-evolved ensemble on the Wine dataset. In a representative co-evolved run, Member 3 is the strongest overall standalone classifier, driven by near-perfect recognition of \texttt{class\_1} and strong performance on \texttt{class\_0}. Member 1 specializes in \texttt{class\_0} and \texttt{class\_2}, while Member 2 provides partial support on \texttt{class\_1} and additional coverage on \texttt{class\_0}. As a result, the three members exhibit complementary class-specific behaviors instead of acting as redundant copies of each other. The ensemble is therefore able to leverage the strengths of each member, yielding substantially higher accuracy than any standalone network. This behavior is consistent with the marginal-contribution fitness formulation, which rewards networks that provide non-redundant contributions to group performance.

\paragraph{B. Regression Tasks:}\mbox{}\\
% original version used for submission:
% The results across all regression datasets, shown in Figure~\ref{fig:regression_results} and summarized in Table~\ref{tab:regression_results}, indicate that co-evolved ensembles consistently achieve lower MAE than both single networks and post-evolved ensembles. The improvement is most pronounced on California Housing and Superconductivity, with smaller but consistent gains on Concrete and Diabetes. Co-evolution also produces more compact error distributions, indicating improved stability across runs, particularly on the Superconductivity dataset where error reduction is accompanied by decreased variance despite the increased nonlinearity of the task. This suggests that cooperative training encourages networks to specialize in complementary regions of the input space, improving ensemble predictions.

% In contrast, post-evolved ensembling yields limited and inconsistent improvements, often failing to meaningfully shift the error distribution. This highlights a limitation of combining independently optimized models, where averaging alone does not ensure complementary behavior. Co-evolved networks, however, are trained to minimize group error directly.
% % , resulting in more accurate and stable predictions.

% updated version:
The results across all regression datasets, shown in Figure~\ref{fig:regression_results} and summarized in Table~\ref{tab:regression_results}, indicate that co-evolved ensembles consistently achieve lower MAE than both baseline methods. The largest improvements are observed on California Housing and Superconductivity, while gains on Diabetes and Concrete are smaller but remain consistent. Co-evolution also produces tighter error distributions, particularly on Superconductivity, indicating improved stability across runs. In contrast, post-evolved ensembling yields only modest and inconsistent improvements, suggesting that averaging independently optimized networks does not reliably produce complementary behavior.

\begin{table}[t]
\caption{Regression Performance (MAE $\downarrow$)}
\label{tab:regression_results}
\centering
\begin{tabular}{lccc}
\toprule
\textbf{Dataset} & \textbf{Single Net.} & \textbf{Post-Evolved} & \textbf{Co-Evolved} \\
 & ($\mu \pm \sigma$) & ($\mu \pm \sigma$) & ($\mu \pm \sigma$) \\
\midrule
Housing & 1.007 $\pm$ 0.223 & 1.020 $\pm$ 0.234 & \textbf{0.887 $\pm$ 0.083} \\
Diabetes & 71.670 $\pm$ 14.679 & 71.611 $\pm$ 14.770 & \textbf{63.562 $\pm$ 9.337} \\
Concrete & 15.769 $\pm$ 3.850 & 15.749 $\pm$ 3.860 & \textbf{14.142 $\pm$ 2.938} \\
Supercond. & 30.132 $\pm$ 3.598 & 29.655 $\pm$ 3.573 & \textbf{28.322 $\pm$ 2.735} \\
\bottomrule
\end{tabular}
\end{table}

\paragraph{C. Control Tasks:}\mbox{}\\
The CartPole-v1 results in Figure~\ref{fig:control_results} exhibit the largest performance gap between methods. Co-evolved ensembles frequently approach the maximum reward, whereas both single networks and post-evolved ensembles remain substantially lower. Although the co-evolved distribution shows greater variance, reflecting the sensitivity of reinforcement learning to small differences in policy behavior, its performance remains consistently stronger across runs. The weakness of post-evolved ensembles highlights the importance of training-time interaction. Unlike classification and regression, where outputs are aggregated directly, control requires compatible action-value estimates at every timestep, and independently evolved networks do not reliably produce compatible policies when combined. These results also illustrate the regime anticipated by the search-space argument of Section~3.5. Whereas co-evolution provides incremental gains on tasks where standard evolution already succeeds, here it enables a qualitative transition from near-failure to near-optimal performance, indicating that its benefits grow with the difficulty of the underlying search problem.

\paragraph{D. Scaling Studies:} \mbox{}\\
%  original version used from submission:
% The scaling results for both the Wine and Superconductivity datasets (Figure~\ref{fig:wine_coevolved_scaling} and Figure~\ref{fig:superconductivity_coevolved_scaling}) show that increasing the size of co-evolved ensembles leads to an initial improvement in performance followed by diminishing returns. For Wine, accuracy improves quickly from two to around five populations before stabilizing, while for Superconductivity, MAE decreases with additional populations but at a progressively slower rate. This consistent pattern across both classification and regression tasks suggests that a relatively small number of interacting populations is sufficient to capture most of the benefits of co-evolution on these tasks. This behavior, however, may not hold for more complex tasks, which may benefit from larger ensembles. 
% % We discuss this more in the next section on Limitations.
% % Once sufficient diversity and complementary behavior are established, additional populations provide limited gains while increasing computational cost, indicating that the approach achieves strong performance without requiring large ensemble sizes even in more complex settings.

% updated version:
The scaling results for both the Wine and Superconductivity datasets (Figures~\ref{fig:wine_coevolved_scaling} and~\ref{fig:superconductivity_coevolved_scaling}) show that increasing ensemble size produces initial performance improvements followed by diminishing returns. Wine accuracy improves rapidly from two to about five populations before stabilizing, while Superconductivity MAE decreases more gradually with additional populations. These results suggest that relatively small ensembles capture most of the benefits of co-evolution, although larger and more complex tasks may require additional populations.

\subsection{Limitations}

We emphasize that this work targets the scale and complexity regime of ultra-low-power neuromorphic hardware such as $\mu$Caspian, where networks are constrained to very small architectures. We do not claim that co-evolution scales to arbitrarily large networks or frontier machine learning tasks. Rather, the claim is that cooperative decomposition enables more difficult problems to be addressed within these hardware constraints. Evaluating the framework on larger neuromorphic platforms and correspondingly more challenging tasks remains an important direction for future work. 

While co-evolved ensembles consistently outperform both baselines, the magnitude of improvement varies. On some datasets, particularly Concrete and Diabetes, the gains over post-hoc ensembling are modest, suggesting that cooperative training is most beneficial when independently evolved networks do not already provide sufficient diversity. 

Deploying an ensemble also introduces inference-time costs. An ensemble of $N$ networks requires roughly $N$ times the computation and memory of a single network. On resource-constrained hardware, this may require sequential execution of the sub-networks, increasing inference latency. The benefits of co-evolution must therefore be considered alongside the increased inference cost.

The expressivity constraint identified in the search space analysis represents another limitation. Because sub-networks do not communicate during inference, the ensemble can only represent functions decomposable as $g(f_1(x),\ldots,f_N(x))$ under the fixed aggregation $g$. Tasks requiring tightly coupled dynamics across many neurons may therefore favor monolithic networks despite the more difficult search problem. 

Finally, the current framework fixes the ensemble size $N$ before training. Although the scaling studies suggest diminishing returns beyond moderate $N$, the optimal ensemble size remains task dependent and should be determined empirically.

% \subsection{Future work}

\section{Conclusions}
In this work, we introduced a co-evolved ensemble framework for SNNs in which fitness is defined by each network's marginal contribution to group performance. By integrating ensemble formation directly into the evolutionary process, the proposed approach encourages complementary specialization during training rather than relying on post-hoc combination of independently optimized networks. Across classification, regression, and reinforcement learning control tasks, co-evolved ensembles consistently improve performance and stability relative to both single-network evolution and post-hoc ensembling.

These results suggest that cooperative decomposition provides a promising approach for addressing the growing complexity of evolutionary optimization. By decomposing a task into smaller interacting sub-networks, the proposed framework reduces the effective search space while preserving complex behavior through cooperative learning. Although demonstrated using EONS, the approach is applicable to a broad class of evolutionary algorithms. Overall, these findings highlight co-evolution as a practical pathway for advancing evolutionary training of SNNs and neuromorphic systems.

\bibliographystyle{ACM-Reference-Format}
\bibliography{references}

\end{document}